\documentclass[11pt,a4paper]{article}
\usepackage[hyperref]{acl2020}
\usepackage{times}
\usepackage{latexsym}

\usepackage{multirow}
\usepackage{}
\usepackage{microtype}

\usepackage{algorithmicx}  
\usepackage{algpseudocode}  
\usepackage{amsmath}  
\usepackage{color,xcolor}
\usepackage{algorithm} 

\usepackage{pifont}
\usepackage{bm}
\usepackage{amsfonts}

\usepackage{graphicx}
\usepackage[normalem]{ulem}
\usepackage{multirow}
\usepackage{amsmath}
\usepackage{url}

\usepackage{amssymb}
\usepackage{amsopn}
\usepackage{subcaption}
\usepackage{graphicx}
\usepackage{multirow}
\usepackage[english]{babel}
\usepackage{bm}
\usepackage{array}

\aclfinalcopy 

\title{A Novel Graph-based Multi-modal Fusion Encoder \\ for Neural Machine Translation}

\author{Yongjing Yin$^{1}\thanks{\ \ This work is done when Yongjing Yin was interning at Pattern Recognition Center, WeChat AI, Tencent Inc, China.}$, \ Fandong Meng$^{2}$, \  Jinsong Su$^{1}\thanks{\ \ Corresponding author.}$, \
Chulun Zhou$^{1}$, \\ {\bf Zhengyuan Yang}$^{3}$, \ {\bf Jie Zhou}$^{2}$, {\bf Jiebo Luo}$^{3}$\\
$^{1}$Xiamen University, Xiamen, China \\ $^{2}$Pattern Recognition Center, WeChat AI, Tencent Inc, Beijing, China \\
$^{3}$Department of Computer Science, University of Rochester, Rochester NY 14627, USA\\
 {\tt yinyongjing@stu.xmu.edu.cn  fandongmeng@tencent.com} \\ {\tt jssu@xmu.edu.cn } \\ 
}

\date{}

\begin{document}
\maketitle

\begin{abstract}
Multi-modal neural machine translation (NMT) aims to translate source sentences into a target language paired with images. However, dominant multi-modal NMT models do not fully exploit fine-grained semantic correspondences between semantic units of different modalities, which have potential to refine multi-modal representation learning.
To deal with this issue, in this paper, we propose a novel graph-based multi-modal fusion encoder for NMT. Specifically, we first represent the input sentence and image using a unified multi-modal graph,
which captures various semantic relationships between multi-modal semantic units (words and visual objects).
We then stack multiple graph-based multi-modal fusion layers that iteratively perform semantic interactions to learn node representations.
Finally, these representations provide an attention-based context vector for the decoder.
We evaluate our proposed encoder on the Multi30K datasets.
Experimental results and in-depth analysis show the superiority of our multi-modal NMT model.
\end{abstract}

\section{Introduction}
Multi-modal neural machine translation (NMT) \cite{Huang:WMT16,Calixto:ACL17} has become an important research direction in machine translation, due to its research significance in multi-modal deep learning and wide applications, such as translating multimedia news and web product information \cite{Zhouemnlp18}.
It significantly extends the conventional text-based machine translation by taking images as additional inputs. 
The assumption behind this is that the translation is expected to be more accurate compared to purely text-based translation, since the visual context helps to resolve ambiguous multi-sense words \cite{acl19:twopass}.


\begin{figure*}
\centering
\includegraphics[width=0.9\linewidth]{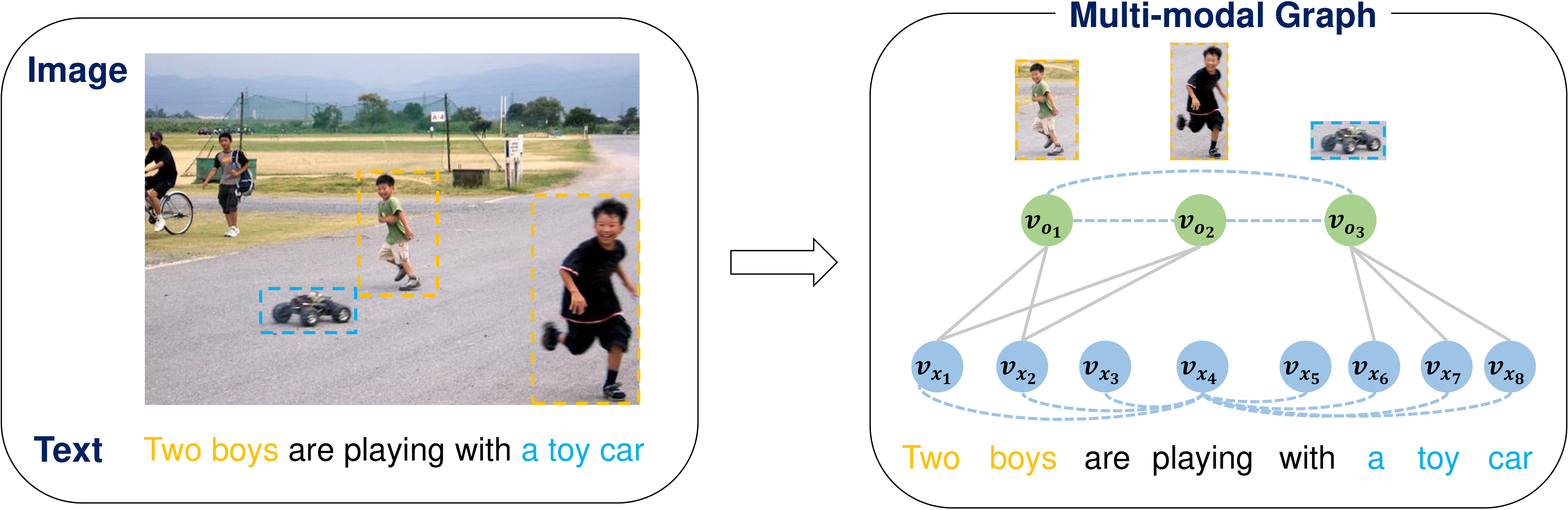}
\caption{
\label{fig:mm-graph}
The multi-modal graph for an input sentence-image pair. The blue and green solid circles denote textual nodes and visual nodes respectively. An intra-modal edge (dotted line) connects two nodes in the same modality, and an inter-modal edge (solid line) links two nodes in different modalities.
Note that we only display edges connecting the textual node ``\emph{playing}'' and other textual ones for simplicity.
}
\end{figure*}

Apparently,
how to fully exploit visual information is one of the core issues in multi-modal NMT,
which directly impacts the model performance.
To this end, a lot of efforts have been made, roughly consisting of:
(1) encoding each input image into a global feature vector, 
which can be used to initialize different components of multi-modal NMT models,
or as additional source tokens \cite{Huang:WMT16,Calixto:ACL17}, 
or to learn the joint multi-modal representation \cite{Zhouemnlp18,vaeacl19};
(2) extracting object-based image features to initialize the model, or supplement source sequences, or generate attention-based visual context \cite{Huang:WMT16,acl19:twopass};
and (3) representing each image as spatial features, 
which can be exploited as extra context \cite{Calixto:ACL17,Delbrouck:EMNLP17,acl19:twopass}, or a supplement to source semantics \cite{Delbrouck:NIPS17workshop} via an attention mechanism.


Despite their success, the above studies do not fully exploit the fine-grained semantic correspondences between semantic units within an input sentence-image pair. For example, as shown in Figure \ref{fig:mm-graph}, the noun phrase ``\emph{a toy car}" semantically corresponds to the blue dashed region.
The neglect of this important clue may be due to two big challenges:
1) how to construct a unified representation to bridge the semantic gap between two different modalities, 
and 2) how to achieve semantic interactions based on the unified representation.
However, we believe that such semantic correspondences can be exploited to refine multi-modal representation learning, since they enable the representations within one modality to incorporate cross-modal information as supplement during multi-modal semantic interactions \cite{stackcross,LXMERT}.

In this paper, 
we propose a novel graph-based multi-modal fusion encoder for NMT. 
We first represent the input sentence and image with a unified multi-modal graph.
In this graph, each node indicates a semantic unit: \emph{textual word} or \emph{visual object}, and two types of edges are introduced to model semantic relationships between semantic units within the same modality (\emph{intra-modal edges}) and semantic correspondences between semantic units of different modalities (\emph{inter-modal edges}) respectively.
Based on the graph, we then stack multiple graph-based multi-modal fusion layers that iteratively perform semantic interactions among the nodes to conduct graph encoding.
Particularly, during this process, we distinguish the parameters of two modalities, 
and sequentially conduct intra- and inter-modal fusions to learn multi-modal node representations.
Finally, these representations can be exploited by the decoder via an attention mechanism.

Compared with previous models, ours is able to fully exploit semantic interactions among multi-modal semantic units for NMT.
Overall, 
the major contributions of our work are listed as follows:
\begin{itemize}
\setlength{\itemsep}{2pt}
\setlength{\parsep}{2pt}
\setlength{\parskip}{2pt}
\item 
We propose a unified graph to represent the input sentence and image, 
where various semantic relationships between multi-modal semantic units can be captured for NMT.

\item We propose a graph-based multi-modal fusion encoder to conduct graph encoding based on the above graph. To the best of our knowledge, our work is the first attempt to explore multi-modal graph neural network (GNN) for NMT.

\item 
We conduct extensive experiments on Multi30k datasets of two language pairs. 
Experimental results and in-depth analysis indicate that our encoder is effective to fuse multi-modal information for NMT. Particularly, our multi-modal NMT model significantly outperforms several competitive baselines. 
\item
We release the code at https://github.com/\\DeepLearnXMU/GMNMT.
\end{itemize}

\section{NMT with Graph-based Multi-modal Fusion Encoder}
Our multi-modal NMT model is based on attentional encoder-decoder framework with maximizing the log likelihood of training data as the objective function.

\begin{figure*}
\centering
\includegraphics[width=0.95\linewidth]{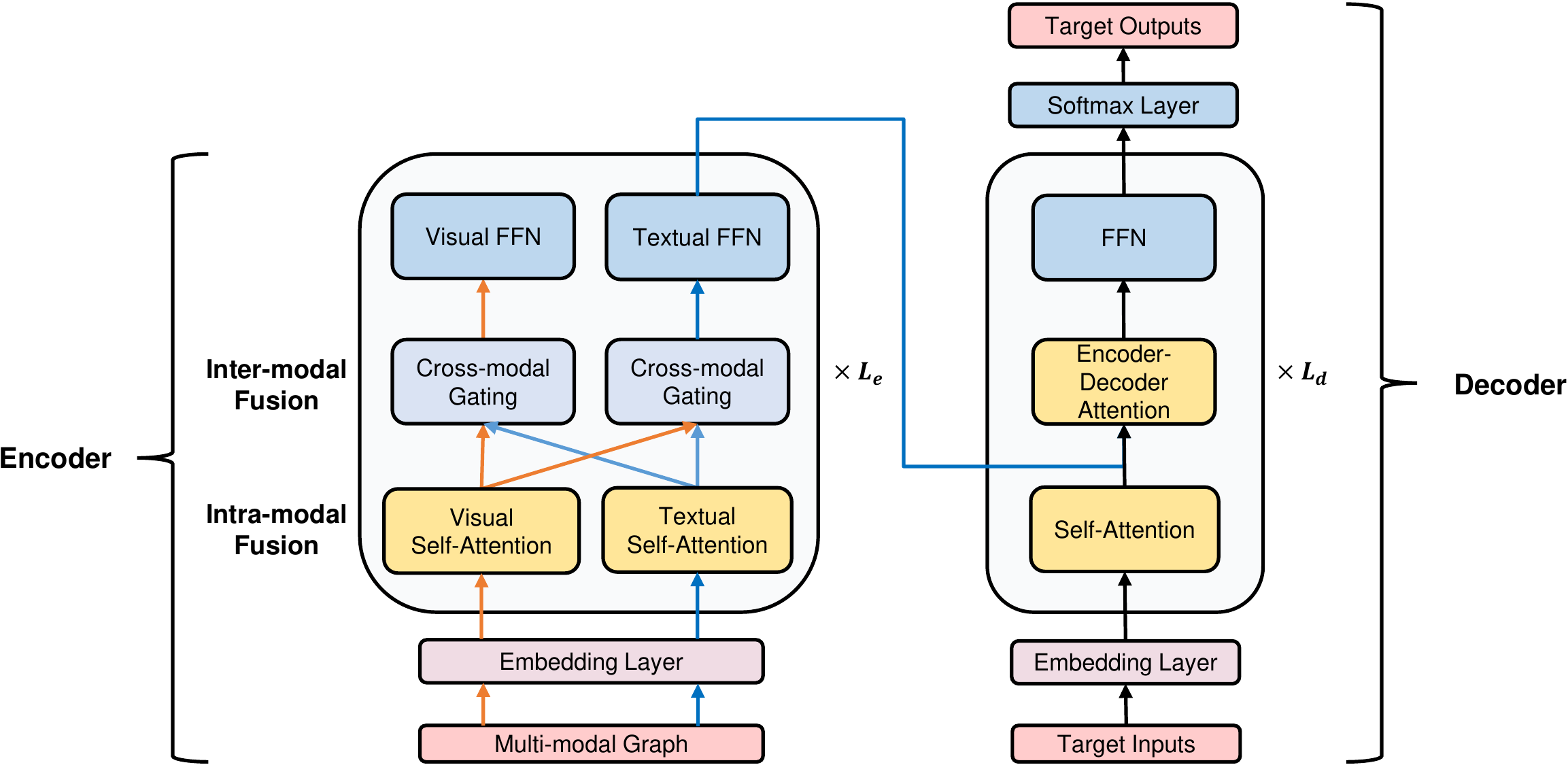}
\caption{
\label{fig:se-graph}
The architecture of our NMT model with the graph-based multi-modal fusion encoder. Note that we actually do not apply a Visual FFN to the last layer in the encoder.}
\end{figure*}

\subsection{Encoder}
Essentially, 
our encoder can be regarded as a multi-modal extension of GNN.
To construct our encoder,
we first represent the input sentence-image pair as a unified multi-modal graph.
Then, based on this graph, 
we stack multiple multi-modal fusion layers to learn node representations, which provides the attention-based context vector to the decoder. 

\subsubsection{Multi-modal Graph}\label{graph}
In this section, we take the sentence and the image shown in Figure \ref{fig:mm-graph} as an example, and describe how to use a multi-modal graph to represent them. 
Formally, our graph is undirected and can be formalized as $G$=($V$,$E$),
which is constructed as follows:

In the node set $V$,
each node represents either a textual word or a visual object.
Specifically,
we adopt the following strategies to construct these two kinds of nodes:
(1) We include all words as separate \textbf{textual nodes} in order to fully exploit textual information.
For example, 
in Figure \ref{fig:mm-graph}, 
the multi-modal graph contains totally eight textual nodes, 
each of which corresponds to a word in the input sentence;
(2) We employ the Stanford parser to identify all noun phrases in the input sentence, and then apply a visual grounding toolkit \cite{yang2019fast} to detect bounding boxes (visual objects) for each noun phrase. 
Subsequently, all detected visual objects are included as independent \textbf{visual nodes}.
In this way, we can effectively reduce the negative impact of abundant unrelated visual objects.
Let us revisit the example in Figure \ref{fig:mm-graph},
where we can identify two noun phrases ``\emph{Two boys}'' and ``\emph{a toy car}'' from the input sentence, and then include three visual objects into the multi-modal graph.

To capture various semantic relationships between multi-modal semantic units for NMT, we consider two kinds of edges in the edge set $E$:
(1) Any two nodes in the same modality are connected by an \textbf{intra-modal edge};
and
(2) Each textual node representing any noun phrase and the corresponding visual node are connected by an \textbf{inter-modal edge}.
Back to Figure \ref{fig:mm-graph},
we can observe that all visual nodes are connected to each other, and all textual nodes are fully-connected. However, only nodes $v_{o_1}$ and $v_{x_1}$, $v_{o_1}$ and $v_{x_2}$, $v_{o_2}$ and $v_{x_1}$, $v_{o_2}$ and $v_{x_2}$, $v_{o_3}$ and $v_{x_6}$, $v_{o_3}$ and $v_{x_7}$, $v_{o_3}$ and $v_{x_8}$ are connected by inter-modal edges.

\subsubsection{Embedding Layer}
Before inputting the multi-modal graph into the stacked fusion layers,
we introduce an embedding layer to initialize the node states.
Specifically, for each textual node $v_{x_i}$,
we define its initial state $H^{(0)}_{x_i}$ as the sum of its word embedding and position encoding \cite{Vaswani:nips17}.
To obtain the initial state $H^{(0)}_{o_j}$ of the visual node $v_{o_j}$,
we first extract visual features from the fully-connected layer that follows the ROI pooling layer in Faster-RCNN \cite{fasterrcnn}, 
and then employ a multi-layer perceptron with ReLU activation function to project these features onto the same space as textual representations.

\subsubsection{Graph-based Multi-modal Fusion Layers}
As shown in the left part of Figure \ref{fig:se-graph}, on the top of embedding layer, we stack $L_e$ \emph{graph-based multi-modal fusion layers} to encode the above-mentioned multi-modal graph.
At each fusion layer, we sequentially conduct intra- and inter-modal fusions to update all node states.
In this way, the final node states encode both the context within the same modality and the cross-modal semantic information simultaneously.
Particularly, 
since visual nodes and textual nodes are two types of semantic units containing the information of different modalities, 
we apply similar operations but with different parameters to model their state update process, respectively.

Specifically, in the $l$-th fusion layer, 
both updates of textual node states $\textbf{H}^{(l)}_x$=$\{H^{(l)}_{x_i}\}$ and visual node states $\textbf{H}^{(l)}_o$=$\{H^{(l)}_{o_j}\}$ mainly involve the following steps:

\textbf{Step1: Intra-modal fusion}.
At this step, we employ \textbf{self-attention} to generate the contextual representation of each node by collecting the message from its neighbors of the same modality.

Formally, the contextual representations $\textbf{C}^{(l)}_x$ of all textual nodes are calculated as follows: \footnote{For simplicity, we omit the descriptions of layer normalization and residual connection.}
\begin{align}
\textbf{C}^{(l)}_x=\text{MultiHead}(\textbf{H}^{(l-1)}_x, \textbf{H}^{(l-1)}_x, \textbf{H}^{(l-1)}_x),
\end{align}
where MultiHead(\textbf{Q}, \textbf{K}, \textbf{V}) is a multi-head self-attention function taking a query matrix \textbf{Q}, a key matrix \textbf{K}, and a value matrix \textbf{V} as inputs. 
Similarly, we generate the contextual representations $\textbf{C}^{(l)}_o$ of all visual nodes as
\begin{equation}
\textbf{C}^{(l)}_o = \text{MultiHead}(\textbf{H}^{(l-1)}_o, \textbf{H}^{(l-1)}_o, \textbf{H}^{(l-1)}_o).
\end{equation}
In particular, since the initial representations of visual objects are extracted from deep CNNs, we apply a simplified multi-head self-attention to preserve the initial representations of visual objects,
where the learned linear projects of values and final outputs are removed.

\textbf{Step2: Inter-modal fusion}. 
Inspired by studies in multi-modal feature fusion \cite{vqatips_cvpr18,bilatt_nips18}, we apply a \textbf{cross-modal gating} mechanism with an element-wise operation to gather the semantic information of the cross-modal neighbours of each node.

Concretely, 
we generate the representation $M^{(l)}_{x_i}$ of a text node $v_{x_i}$ in the following way:
\begin{align}
M^{(l)}_{x_i} &= \sum_{j\in A(v_{x_i})} \alpha_{i,j} \odot C^{(l)}_{o_j},\\
\alpha_{i,j}  &= \text{Sigmoid}(\textbf{W}^{(l)}_1C^{(l)}_{x_i}+\textbf{W}^{(l)}_2C^{(l)}_{o_j}),
\end{align}
where $A(v_{x_i})$ is the set of neighboring visual nodes of $v_{x_i}$, and $\textbf{W}^{(l)}_1$ and $\textbf{W}^{(l)}_2$ are parameter matrices.
Likewise, we produce the representation $M^{(l)}_{o_j}$ of a visual node $v_{o_j}$ as follows:
\begin{align}
M^{(l)}_{o_j}&=\sum_{i\in A(v_{o_j})} \beta_{j,i} \odot C^{(l)}_{x_i},\\
\beta_{j,i}&=\text{Sigmoid}(\textbf{W}^{(l)}_3C^{(l)}_{o_j}+\textbf{W}^{(l)}_4C^{(l)}_{x_i}),
\end{align}
where $A(v_{o_j})$ is the set of adjacent textual nodes of $v_{o_j}$, and $\textbf{W}^{(l)}_3$ and $\textbf{W}^{(l)}_4$ are also parameter matrices.

The advantage is that the above fusion approach can better determine the degree of inter-modal fusion according to the contextual representations of each modality.
Finally, 
we adopt position-wise feed forward networks $\text{FFN}(*)$ to generate the textual node states $\textbf{H}_x^{(l)}$ and visual node states $\textbf{H}_o^{(l)}$:
\begin{align}
\textbf{H}_x^{(l)}=\text{FFN}(\textbf{M}^{(l)}_x),\\
\textbf{H}_o^{(l)}=\text{FFN}(\textbf{M}^{(l)}_o),
\end{align}
where $\textbf{M}^{(l)}_x$ = $\{M^{(l)}_{x_i}\}$, $\textbf{M}^{(l)}_o$ = $\{M^{(l)}_{o_j}\}$ denote the above updated representations of all textual nodes and visual nodes respectively.

\subsection{Decoder}\label{decoder}
Our decoder is similar to the conventional Transformer decoder. 
Since visual information has been incorporated into all textual nodes via multiple graph-based multi-modal fusion layers, 
we allow the decoder to dynamically exploit the multi-modal context by only attending to textual node states.

As shown in the right part of Figure \ref{fig:se-graph}, 
we follow \citet{Vaswani:nips17} to stack $L_d$ identical layers to generate target-side hidden states,
where each layer $l$ is composed of three sub-layers.
Concretely, the first two sub-layers are a masked self-attention and an encoder-decoder attention to integrate target- and source-side contexts respectively:
\begin{align}
\textbf{E}^{(l)} &= \text{MultiHead}(\textbf{S}^{(l-1)},\textbf{S}^{(l-1)},\textbf{S}^{(l-1)}),\\
\textbf{T}^{(l)} &= \text{MultiHead}(\textbf{E}^{(l)}, \textbf{H}^{(L_e)}_x, \textbf{H}^{(L_e)}_x),
\end{align}
where
$\textbf{S}^{(l-1)}$ denotes the target-side hidden states in the $l$-$1$-th layer.
In particular, 
$\textbf{S}^{(0)}$ are the embeddings of input target words.
Then, a position-wise fully-connected forward neural network is uesd to produce $\textbf{S}^{(l)}$ as follows:
\begin{align}
\textbf{S}^{(l)} = \text{FFN}(\textbf{T}^{(l)}).
\end{align}

Finally, the probability distribution of generating the target sentence is defined by using a softmax layer, which takes the hidden states in the top layer as input:
\begin{align}
P(Y|X,I)=\prod_{t} \text{Softmax}(\textbf{W}\textbf{S}^{(L_d)}_t+b),
\end{align}
where $X$ is the input sentence, $I$ is the input image, $Y$ is the target sentence, and $\textbf{W}$ and $b$ are the parameters of the softmax layer.


\section{Experiment}
We carry out experiments on multi-modal English$\Rightarrow$German (En$\Rightarrow$De) and English$\Rightarrow$French (En$\Rightarrow$Fr) translation tasks.

\subsection{Setup}
\paragraph{Datasets} 
We use the Multi30K dataset \cite{Multi30K}, where each image is paired with one English description and human translations into German and French.
Training, validation and test sets contain 29,000, 1,014 and 1,000 instances respectively. 
In addition, we evaluate various models on the WMT17 test set and the ambiguous MSCOCO test set, which contain 1,000 and 461 instances respectively.
Here,  we directly use the preprocessed sentences \footnote{http://www.statmt.org/wmt18/multimodal-task.html} and segment words into subwords via byte pair encoding \cite{bpe:ACL2016} with 10,000 merge operations. 

\paragraph{Visual Features}
We first apply the Stanford parser to identify noun phrases from each source sentence, 
and then employ the visual ground toolkit released by \citet{yang2019fast} to detect associated visual objects of the identified noun phrases.
For each phrase, we keep the visual object with the highest prediction probability,
so as to reduce negative effects of abundant visual objects. 
In each sentence, the average numbers of objects and words are around 3.5 and 15.0 respectively. \footnote{There is no parsing failure for this dataset. If no noun is detected for a sentence, the object representations will be set to zero vectors and the model will degenerate to Transformer.}
Finally, we compute 2,048-dimensional features for these objects with the pre-trained ResNet-100 Faster-RCNN \cite{fasterrcnn}.

\paragraph{Settings}
We use Transformer \cite{Vaswani:nips17} as our baseline. 
Since the size of training corpus is small and the trained model tends to be over-fitting, 
we first perform a small grid search to obtain a set of hyper-parameters on the En$\Rightarrow$De validation set. 
Specifically, 
the word embedding dimension and hidden size are 128 and 256 respectively. 
The decoder has $L_d$=4 layers\footnote{The encoder of the text-based Transformer also has 4 layers.}
and the number of attention heads is 4. 
The dropout is set to 0.5. Each batch consists of approximately 2,000 source and target tokens.
We apply the Adam optimizer with a scheduled learning rate to optimize various models, and we use other same settings as \cite{Vaswani:nips17}.
Finally, 
we use the metrics BLEU \cite{bleu} and METEOR \cite{meteor} to evaluate the quality of translations.
Particularly, we run all models three times for each experiment and report the average results.

\paragraph{Baseline Models}
In addition to the text-based Transformer \cite{Vaswani:nips17},
we adapt several effective approaches to Transformer using our visual features, and compare our model with them\footnote{We use suffixes ``(\textbf{RNN})'' and ``(\textbf{TF})'' to represent RNN- and Transformer-style NMT models, respectively.}:
\begin{itemize}
\setlength{\itemsep}{0pt}
\setlength{\parsep}{0pt}
\setlength{\parskip}{0pt}
\item 
\textbf{ObjectAsToken(TF)} \cite{Huang:WMT16}.
It is a variant of the Transformer, where all visual objects are regarded as extra source tokens and placed at the front of the input sentence.
\item 
\textbf{Enc-att(TF)} \cite{Delbrouck:NIPS17workshop}. 
An encoder-based image attention mechanism is incorporated into Transformer, 
which augments each source annotation with an attention-based visual feature vector.

\item
\textbf{Doubly-att(TF)} \cite{Helclwmt18}. 
It is a doubly attentive Transformer.
In each decoder layer, a cross-modal multi-head attention sub-layer is inserted before the fully connected feed-forward layer to generate the visual context vector from visual features.

\end{itemize}

\begin{figure}[!t]
\centering
\includegraphics[width=0.9\linewidth]{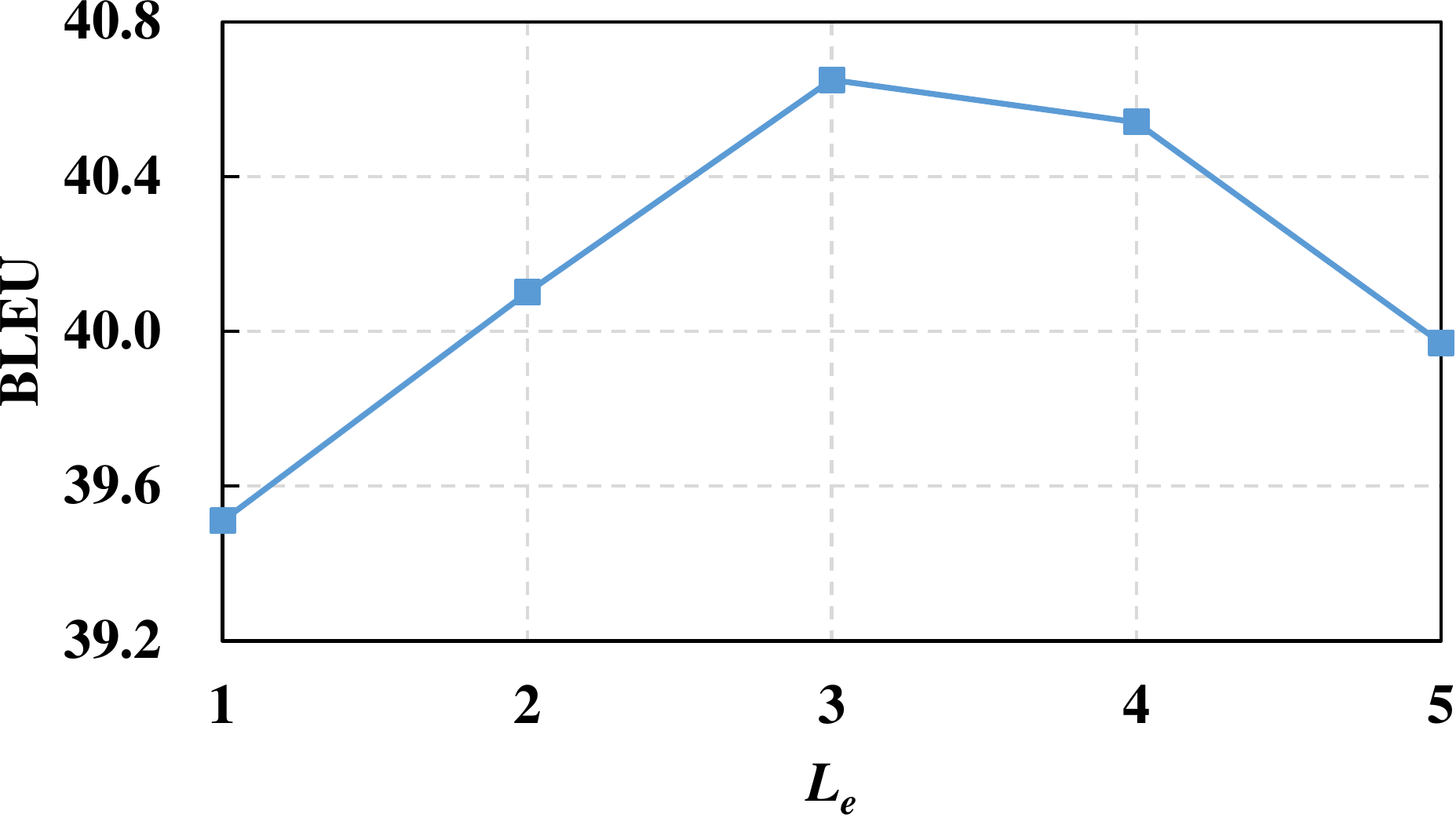}
\caption{
\label{dev}
Results on the En$\Rightarrow$De validation set regarding the number $L_e$ of graph-based multi-modal fusion layers.}
\end{figure}

\begin{table*}[!t]
\centering
\small
\linespread{1.2}
\setlength{\tabcolsep}{1.5mm}{
\begin{tabular}{l|cccccc}
\hline
\multirow{3}{*}{\bf Model} & \multicolumn{6}{c}{\textbf{En$\Rightarrow$De}}  \\
\cline{2-7}
& \multicolumn{2}{c}{\textbf{Test2016}}  
& \multicolumn{2}{c}{\textbf{Test2017}} 
& \multicolumn{2}{c}{\textbf{MSCOCO}} \\
\cline{2-7}
& BLEU & METEOR & BLEU & METEOR & BLEU & METEOR \\
\hline
\hline
\multicolumn{7}{c}{\emph{Existing Multi-modal NMT Systems}} \\
\hline
Doubly-att(RNN) \cite{Calixto:ACL17}  & 36.5 & 55.0 & - & - & - & -  \\
Soft-att(RNN) \cite{Delbrouck:EMNLP17}  & 37.6 & 55.3 & - & - & - & -  \\
Stochastic-att(RNN) \cite{Delbrouck:EMNLP17} & 38.2 & 55.4 & - & - & - & -  \\
Fusion-conv(RNN) \cite{wmt17} & 37.0 & 57.0 & 29.8 & 51.2 & 25.1 & 46.0 \\
Trg-mul(RNN)\cite{wmt17} & 37.8 & \textbf{57.7} & 30.7 & \textbf{52.2} & 26.4 & 47.4 \\
VMMT(RNN) \cite{vaeacl19} & 37.7 & 56.0 & 30.1 & 49.9 & 25.5 & 44.8 \\
Deliberation Network(TF) \cite{acl19:twopass} & 38.0 & 55.6 & - & - & - & - \\
\hline
\multicolumn{7}{c}{\emph{Our Multi-modal NMT Systems}} \\
\hline
Transformer \cite{Vaswani:nips17} & 38.4 & 56.5 & 30.6 & 50.4 & 27.3 & 46.2 \\
ObjectAsToken(TF) \cite{Huang:WMT16} & 39.0 & 57.2 & 31.7 & 51.3 & 28.4 & 47.0 \\
Enc-att(TF) \cite{Delbrouck:NIPS17workshop} & 38.7 & 56.6 & 31.3 & 50.6 & 28.0 & 46.6 \\
Doubly-att(TF) \cite{Helclwmt18} & 38.8 & 56.8 & 31.4 & 50.5 & 27.4 & 46.5 \\	
\hline
Our model & \textbf{39.8} & 57.6 & \textbf{32.2} & 51.9 & \textbf{28.7} & \textbf{47.6} \\
\hline
\end{tabular}}
\caption{
\label{Table_En2DeMainResults}
Experimental results on the En$\Rightarrow$De translation task. 
}
\end{table*}

We also display the performance of several dominant multi-modal NMT models 
such as \textbf{Doubly-att(RNN)} \cite{Calixto:ACL17}, \textbf{Soft-att(RNN)} \cite{Delbrouck:EMNLP17}, \textbf{Stochastic-att(RNN)} \cite{Delbrouck:EMNLP17}, \textbf{Fusion-conv(RNN)} \cite{wmt17}, \textbf{Trg-mul(RNN)} \cite{wmt17}, \textbf{VMMT(RNN)} \cite{vaeacl19} and \textbf{Deliberation Network(TF)} \cite{acl19:twopass} on the same datasets.

\subsection{Effect of Graph-based Multi-modal Fusion Layer Number $L_e$}
The number $L_e$ of multi-modal fusion layer is an important hyper-parameter that directly determines the degree of fine-grained semantic fusion in our encoder.
Thus, 
we first inspect its impact on the EN$\Rightarrow$DE validation set. 

Figure \ref{dev} provides the experimental results using different $L_e$ and our model achieves the best performance when $L_e$ is 3.
Hence, we use $L_e$=3 in all subsequent experiments.

\subsection{Results on the En$\Rightarrow$De Translation Task}\label{SubSection_En2DeMainResults}
Table \ref{Table_En2DeMainResults} shows the main results on the En$\Rightarrow$De translation task. 
Ours outperforms most of the existing models and all baselines, and is comparable to Fusion-conv(RNN) and Trg-mul(RNN) on METEOR. The two results are from the state-of-the-art system on the WMT2017 test set, which is selected based on METEOR. 
Comparing the baseline models, we draw the following interesting conclusions: 

\textbf{First}, 
our model outperforms ObjectAsToken(TF), 
which concatenates regional visual features with text to form attendable sequences and employs self-attention mechanism to conduct inter-modal fusion.
The underlying reasons consist of two aspects:
explicitly modeling semantic correspondences between semantic units of different modalities, and distinguishing model parameters for different modalities.

\textbf{Second}, 
our model also significantly outperforms Enc-att(TF).
Note that Enc-att(TF) can be considered as a single-layer semantic fusion encoder.
In addition to the advantage of explicitly modeling semantic correspondences, we conjecture that multi-layer multi-modal semantic interactions are also beneficial to NMT.

\textbf{Third}, 
compared with Doubly-att(TF) simply using an attention mechanism to exploit visual information,
our model achieves a significant improvement, because of sufficient multi-modal fusion in our encoder.

\begin{figure}[!t]
\centering
\includegraphics[width=0.9\linewidth]{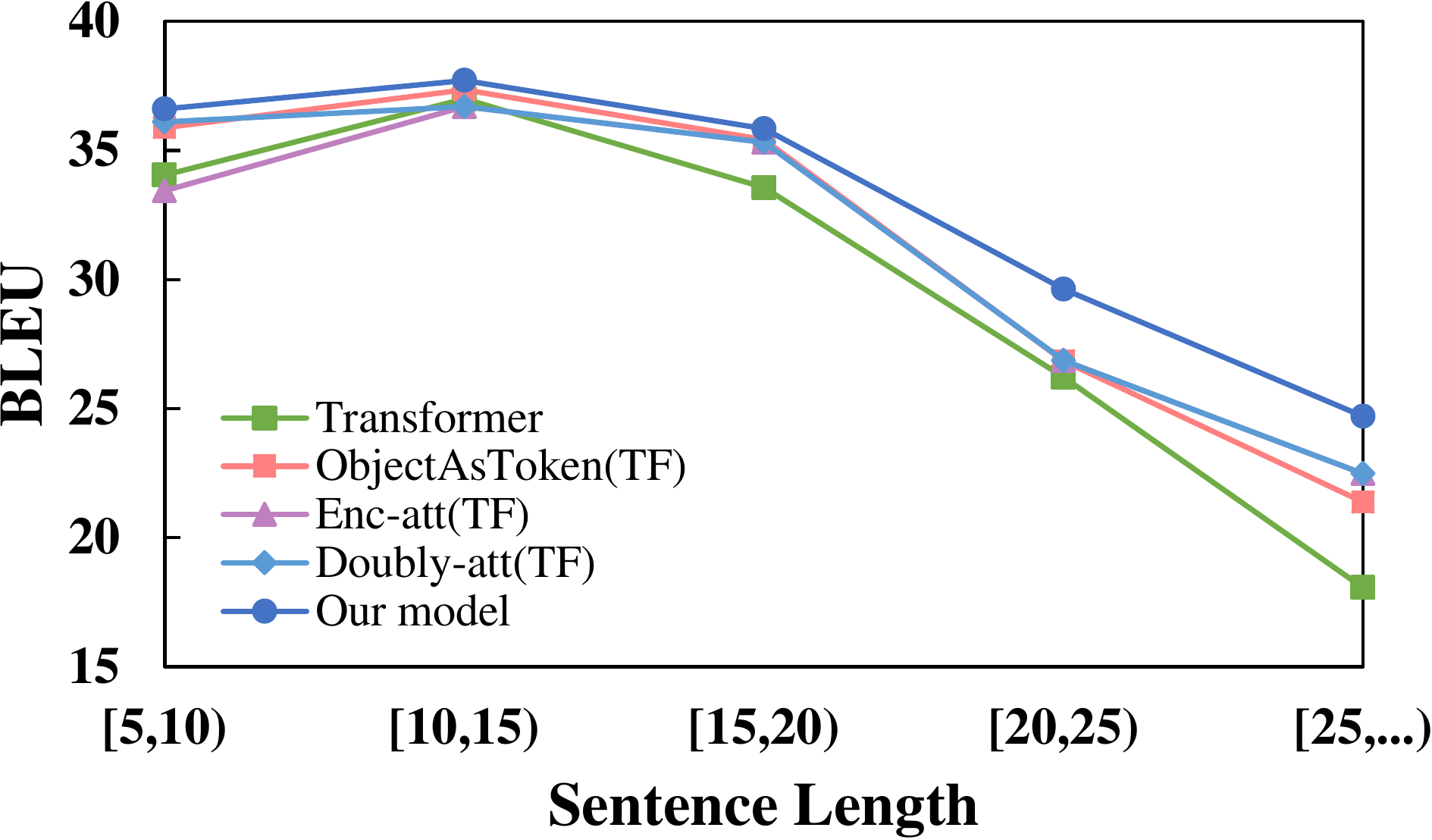}
\caption{
\label{Fig_En2De_LengthEffect}
BLEU scores on different translation groups divided according to source sentence lengths.
}
\end{figure}

\begin{figure}[!t]
\centering
\includegraphics[width=0.9\linewidth]{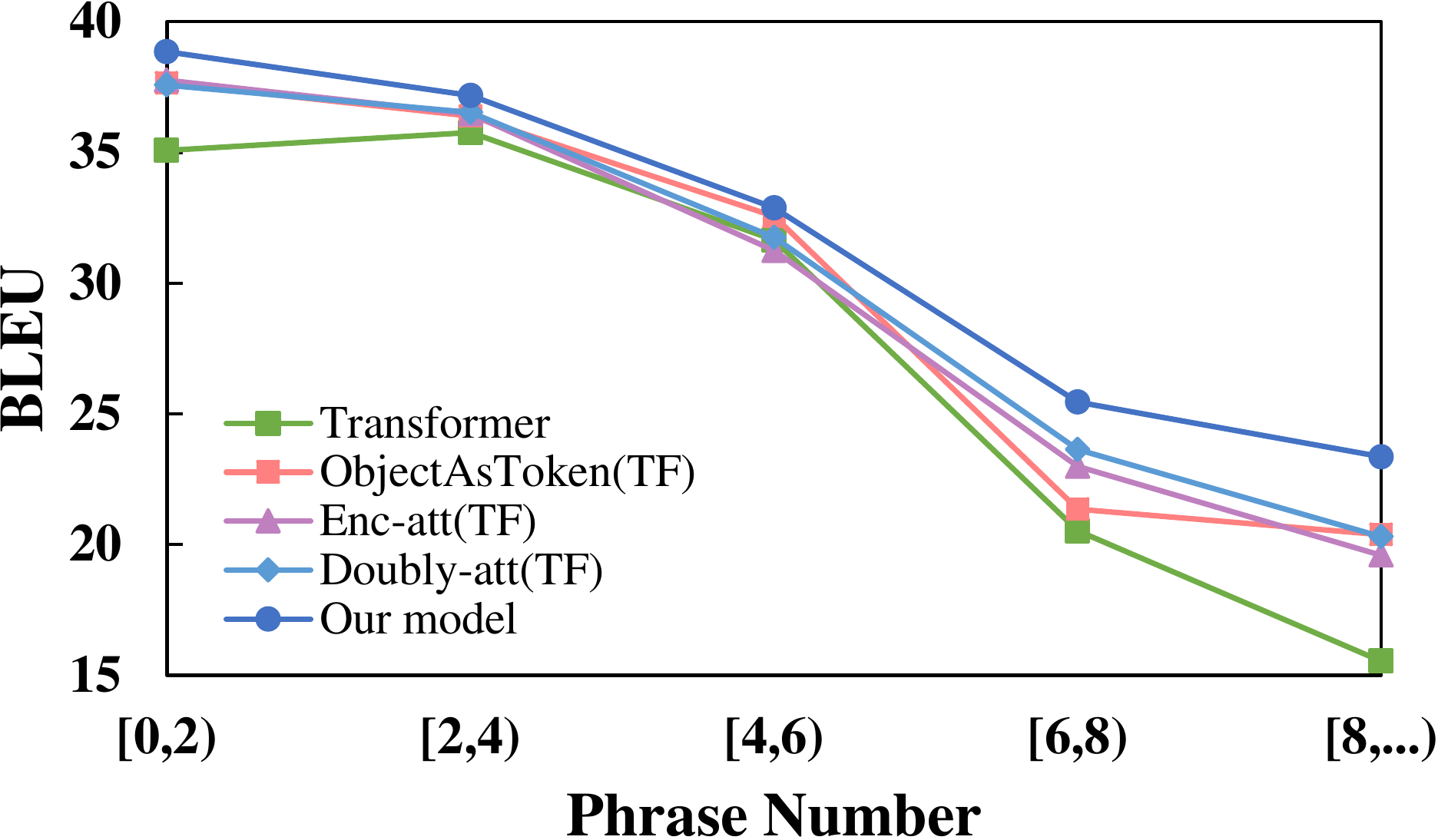}
\caption{
\label{Fig_En2De_PhraseNumberEffect}
BLEU scores on different translation groups divided according to source phrase numbers.
}
\end{figure}

\begin{table*}
\centering
\small
\linespread{1.2}
\setlength{\tabcolsep}{1.2mm}{
\begin{tabular}{l|cccccc}
\hline
\multirow{3}{*}{\bf Model} & \multicolumn{6}{c}{\textbf{En$\Rightarrow$De}} \\ 
\cline{2-7}
& \multicolumn{2}{c}{\textbf{Test2016}} & \multicolumn{2}{c}{\textbf{Test2017}} & \multicolumn{2}{c}{\textbf{MSCOCO}} \\
\cline{2-7}
& BLEU & METEOR & BLEU & METEOR & BLEU & METEOR \\
\hline
\hline
Our model & 39.8 & 57.6 & 32.2 & 51.9 & 28.7 & 47.6 \\
\hline
\ \ w/o inter-modal fusion & 38.7 & 56.7 & 30.7 & 50.6 & 27.0 & 46.7 \\
\ \ visual grounding $\Rightarrow$ fully-connected & 36.4 & 53.4 & 28.3 & 47.0 & 24.4 & 42.9 \\
\hline
\ \ different parameters $\Rightarrow$ unified parameters & 39.2 & 57.3 & 31.9 & 51.4 & 27.7 & 47.4 \\
\hline
\ \ w/ attending to visual nodes & 39.6 & 57.3 & 32.0 & 51.3 & 27.9 & 46.8 \\
\ \ attending to textual nodes $\Rightarrow$ attending to visual nodes & 30.9 & 48.6 & 22.3 & 41.5 & 20.4 & 38.7 \\
\hline
\end{tabular}}
\caption{
\label{Analysis}
Ablation study of our model on the EN$\Rightarrow$DE translation task.
}
\end{table*}

\begin{table*}
\centering
\small
\begin{tabular}{l|cccc}  
\hline
\multirow{3}{*}{\bf Model} & \multicolumn{4}{c}{\textbf{En$\Rightarrow$Fr}}\\
\cline{2-5}
& \multicolumn{2}{c}{\textbf{Test2016}} & \multicolumn{2}{c}{\textbf{Test2017}}\\
\cline{2-5}
& BLEU & METEOR & BLEU & METEOR \\
\hline
\hline
\multicolumn{5}{c}{\emph{Existing Multi-modal NMT Systems}} \\
\hline
Fusion-conv(RNN) \cite{wmt17} & 53.5 & 70.4 & 51.6 & 68.6 \\
Trg-mul(RNN)\cite{wmt17} & 54.7 & 71.3 & 52.7 & \textbf{69.5} \\
Deliberation Network(TF) \cite{acl19:twopass} & 59.8 & 74.4 & - & - \\
\hline
\multicolumn{5}{c}{\emph{Our Multi-modal NMT Systems}} \\
\hline
Transformer \cite{Vaswani:nips17}    & 59.5 & 73.7 & 52.0 & 68.0 \\
ObjectAsToken(TF) \cite{Huang:WMT16} & 60.0 & 74.3 & 52.9 & 68.6 \\	
Enc-att(TF) \cite{Delbrouck:NIPS17workshop} & 60.0 & 74.3 & 52.8 & 68.3 \\
Doubly-att(TF) \cite{Helclwmt18}     & 59.9 & 74.1 & 52.4 & 68.1 \\
\hline
Our model & \textbf{60.9} & \textbf{74.9} & \textbf{53.9} & 69.3 \\
\hline
\end{tabular}
\caption{
\label{Table_En2FrCsMainResults}
Experimental results on the En$\Rightarrow$Fr translation task.
}
\end{table*}

Besides, 
we divide our test sets into different groups based on the lengths of source sentences and the numbers of noun phrases, 
and then compare the performance of different models in each group.
Figures \ref{Fig_En2De_LengthEffect} and \ref{Fig_En2De_PhraseNumberEffect} report the BLEU scores on these groups. 
Overall, our model still consistently achieves the best performance in all groups.
Thus, we confirm again the effectiveness and generality of our proposed model. 
Note that in the sentences with more phrases, which are usually long sentences, the improvements of our model over baselines are more significant.
We speculate that long sentences often contain more ambiguous words. Thus compared with short sentences, long sentences may require visual information to be better exploited as supplementary information, which can be achieved by the multi-modal semantic interaction of our model.

\begin{table}
\centering
\small
\setlength{\tabcolsep}{1.6mm}{
\begin{tabular}{l|ccc}
\hline
\bf Model & Training & Decoding & Parameter \\
\hline
Transformer    & 2.6K & 17.8 & 3.4M  \\
ObjectAsToken(TF) & 1.6K & 17.2 & 3.7M \\	
Enc-att(TF) & 1.3K & 16.9 & 3.6M \\
Doubly-att(TF) & 1.0K & 12.9 & 3.8M \\
Our model & 1.1K & 16.7 & 4.0M \\
\hline
\end{tabular}
}
\caption{
\label{complexity}
Training speed (tokens/second), decoding speed (sentences/second) and the number of parameters of different models on the En$\Rightarrow$De translation task. 
}
\end{table}

We also show the training and decoding speed of our model and the baselines in Table \ref{complexity}. 
During training, our model can process approximately 1.1K tokens per second, which is comparable to other multi-modal baselines. When it comes to decoding procedure, our model translates about 16.7 sentences per second and the speed drops slightly compared to Transformer. Moreover, our model only introduces a small number of extra parameters and achieves better performance.

\subsection{Ablation Study}
To investigate the effectiveness of different components, we further conduct experiments to compare our model with the following variants in Table \ref{Analysis}:

(1) \emph{w/o inter-modal fusion}.
In this variant, 
we apply two separate Transformer encoders to learn the semantic representations of words and visual objects, respectively, and then use the doubly-attentive decoder \cite{Helclwmt18} to incorporate textual and visual contexts into the decoder.
The result in line \textcolor{blue}{3} indicates that removing the inter-modal fusion leads to a significant performance drop. It suggests that semantic interactions among multi-modal semantic units are indeed useful for multi-modal representation learning.

(2) \emph{visual grounding $\Rightarrow$ fully-connected}.
We make the words and visual objects fully-connected to establish the inter-modal correspondences.
The result in line \textcolor{blue}{4} shows that this change causes a significant performance decline.
The underlying reason is the fully-connected semantic correspondences introduce much noise to our model.

(3) \emph{different parameters $\Rightarrow$ unified parameters}.
When constructing this variant, 
we assign unified parameters to update node states in different modalities.
Apparently,
the performance drop reported in line \textcolor{blue}{5} also demonstrates the validity of our approach using different parameters.

(4) \emph{w/ attending to visual nodes}.
Different from our model attending to only textual nodes, 
we allow our decoder of this variant to consider both two types of nodes using doubly-attentive decoder.
From line \textcolor{blue}{6},
we can observe that considering all nodes does not bring further improvement. 
The result confirms our previous assumption that visual information has been fully incorporated into textual nodes in our encoder.

(5) \emph{attending to textual nodes $\Rightarrow$ attending to visual nodes}.
However, when only considering visual nodes, the model performance drops drastically (line \textcolor{blue}{7}).
This is because the number of visual nodes is far fewer than that of textual nodes, which is unable to produce sufficient context for translation.

\begin{figure*}[!t]
\centering
\includegraphics[width=0.85\linewidth]{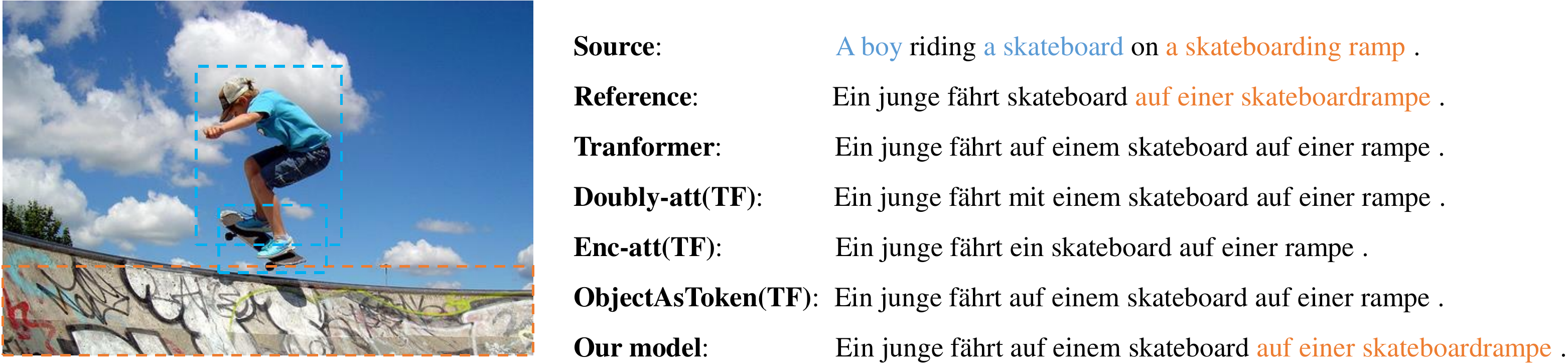}
\caption{
\label{case}
A translation example of different multi-modal NMT models. 
The baseline models do not accurately understand the phrase ``\emph{a skateboarding ramp}" (orange), while our model correctly translate it.
}
\end{figure*}

\subsection{Case Study}
Figure \ref{case} displays the 1-best translations of a sampled test sentence generated by different models.
The phrase ``\emph{a skateboarding ramp}" is not translated correctly by all baselines, while our model correctly translates it. This reveals that our encoder is able to learn more accurate representations.

\subsection{Results on the En$\Rightarrow$Fr Translation Task}
We also conduct experiments on the EN$\Rightarrow$Fr dataset. From Table \ref{Table_En2FrCsMainResults}, our model still achieves better performance compared to all baselines, which demonstrates again that our model is effective and general to different language pairs in multi-modal NMT.

\section{Related Work}
\paragraph{Multi-modal NMT}
\citet{Huang:WMT16} first incorporate global or regional visual features into attention-based NMT.
\citet{Calixto:EMNLP17} also study the effects of incorporating global visual features into different NMT components. 
\citet{imagination} share an encoder between a translation model and an image prediction model to learn visually grounded representations.
Besides, the most common practice is to use attention mechanisms to extract visual contexts for multi-modal NMT \cite{Caglayan:arxiv16,Calixto:ACL17,Delbrouck:EMNLP17,Delbrouck:NIPS17workshop,Findings}.
Recently, 
\citet{acl19:twopass} propose a translate-and-refine approach and 
\citet{vaeacl19} employ a latent variable model to capture the multi-modal interactions for multi-modal NMT.
Apart from model design, \citet{AdvEvaemnlp18} reveal that visual information seems to be ignored by the multi-modal NMT models. \citet{nacl19prob} conduct a systematic analysis and show that visual information can be better leveraged under limited textual context.

Different from the above-mentioned studies, 
we first represent the input sentence-image pair as a unified graph, 
where various semantic relationships between multi-modal semantic units can be effectively captured for multi-modal NMT.
Benefiting from the multi-modal graph, 
we further introduce an extended GNN to conduct graph encoding via multi-modal semantic interactions.

Note that if we directly adapt the approach proposed by \citet{Huang:WMT16} into Transformer, the model (ObjectAsToken(TF)) also involves multi-modal fusion. 
However, ours is different from it in following aspects:
(1) We first learn the contextual representation of each node within the same modality, 
so that it can better determine the degree of inter-modal fusion according to its own context.
(2) We assign different encoding parameters to different modalities, which has been shown effective in our experiments.

Additionally, the recent study LXMERT \cite{LXMERT} also models relationships between vision and language, which differs from ours in following aspects:
(1) \citet{LXMERT} first apply two transformer encoders for two modalities, and then stack two cross-modality encoders to conduct multi-modal fusion. 
In contrast, we sequentially conduct self-attention and cross-modal gating at each layer. 
(2) \citet{LXMERT} leverage an attention mechanism to implicitly establish cross-modal relationships via large-scale pretraining, while we utilize visual grounding to capture explicit cross-modal correspondences.
(3) We focus on multi-modal NMT rather than vision-and-language reasoning in \cite{LXMERT}.

\paragraph{Graph Neural Networks}
Recently, GNNs \cite{Gori:IJCNN05} including gated graph neural network \cite{Li:ICLR16}, graph convolutional network \cite{Duvenaud:NIPS15,Kipf:ICLR17} and graph attention network \cite{gat_iclr18} have been shown effective in many tasks such as VQA \cite{Teney:CVPR17,Brownnips18,Li_2019_ICCV}, text generation \cite{Gildea:ACL18,Becky:ACL18,song2018graph, song2019semantic} and text representation \cite{Zhang:ACL18, Yin, song2018exploring, Xue}.

In this work, we mainly focus on how to extend GNN to fuse multi-modal information in NMT.
Close to our work, 
\citet{Teney:CVPR17} introduce GNN for VQA. 
The main difference between their work and ours is that they build an individual graph for each modality, while we use a unified multi-modal graph.

\section{Conclusion}
In this paper, 
we have proposed a novel graph-based multi-modal fusion encoder, 
which exploits various semantic relationships between multi-modal semantic units for NMT.
Experiment results and analysis on the Multi30K dataset demonstrate the effectiveness of our model.

In the future, 
we plan to incorporate attributes of visual objects and dependency trees to enrich the multi-modal graphs. 
Besides, 
how to introduce scene graphs into multi-modal NMT is a worthy problem to explore. 
Finally, 
we will apply our model into other multi-modal tasks such as multi-modal sentiment analysis.

\section*{Acknowledgments}
This work was supported by the Beijing Advanced Innovation Center for Language Resources (No. TYR17002), the National Natural Science Foundation of China (No. 61672440), and the Scientific Research Project of National Language Committee of China (No. YB135-49).

\bibliography{acl2020}
\bibliographystyle{acl_natbib}

\appendix

\end{document}